\documentclass[12pt]  {iopart}

\usepackage[utf8]{inputenc} 
\usepackage[T1]{fontenc} 

\usepackage{palatino} 
\usepackage{multicol,graphicx}
\usepackage{subcaption}
\usepackage{placeins}
\usepackage{booktabs}
\usepackage{titlesec}
\usepackage{graphicx}
\usepackage{filecontents}
\usepackage{placeins}
\usepackage[export]{adjustbox}
\usepackage{makecell}
\usepackage{color}
\usepackage{float}
\usepackage{enumitem}
\usepackage{labels}
\usepackage{url}

\usepackage[margin=2cm]{geometry}

\usepackage{tabularx}
\usepackage{scrextend}

\usepackage[autostyle=true]{csquotes}
\graphicspath{ {Figures/} }

\begin{document}

\title[Recurrent Neural Networks for P300-based BCI]{Recurrent Neural Networks for P300-based BCI}

\author{Ori Tal and Doron Friedman}

\address{The Advanced Reality Lab, The Interdisciplinary Center, Herzliya, Israel}
\ead{doronf@idc.ac.il}
\vspace{10pt}

\begin{abstract}
P300-based spellers are one of the main methods for EEG-based brain-computer interface, and the detection of the P300 target event with high accuracy is an important prerequisite. The rapid serial visual presentation (RSVP) protocol is of high interest because it can be used by patients who have lost control over their eyes. In this study we wish to explore the suitability of recurrent neural networks (RNNs) as a machine learning method for identifying the P300 signal in RSVP data. We systematically compare RNN with alternative methods such as linear discriminant analysis (LDA) and convolutional neural network (CNN). Our results indicate that LDA performs as well as the neural network models or better on single subject data, but a network combining CNN and RNN has advantages when transferring learning among subejcts, and is significantly more resilient to temporal noise than other methods.
\end{abstract}

%
%
%
%
%

\section{Introduction}

\vspace{0.4cm}

Neural networks have recently been shown to achieve outstanding performance in several machine learning domains such as image recognition \cite{krizhevsky2012imagenet} and voice recognition ~\cite{hinton2012deep}. Most of these breakthroughs have been achieved with convolutional neural networks (CNNs)~\cite{Lenet98}, but some promising results has also been demonstrated by using recurrent neural networks (RNNs) for tasks such as speech and handwriting recognition~\cite{graves2013speech, graves2008unconstrained}, usually when using the long short-term memory (LSTM) architecture ~\cite{LSTM_origin}.

There have been some studies on using ``deep neural networks'' for P300 classification \cite{P300_CNN, RSVP_P300_geva}. The results reported, despite some success, do not show the same dramatic progress achieved by `deep learning' methods as compared to the previous state of the art; while in areas such as image or voice recognition `deep' neural networks have resulted in classification accuracy exceeding other methods by far, this has not yet been the case with EEG in general and P300 detection specifically. The small number of samples typically available in neuroscience (or BCI) is most likely one of the main reasons. In addition, the high dimensionality of the EEG signal, the low signal to noise (SNR) and the existence of outliers in the data, pose other difficulties when trying to use neural networks for BCI tasks (see \cite{lotte2007review}). The main question in this research is whether the RNN model, and particularly LSTM, can enhance the accuracy of P300-based BCI systems and if so, under what conditions.

\vspace{0.4cm}
\section{Background}
\vspace{0.4cm}

P300-based BCI systems can identify when a subject's attention is distracted toward a target event by examining the subject's electroencephalogram (EEG) data. The first system that used the P300 effect was presented by~\cite{FirstP300} and since then different versions of P300 based BCI systems were suggested. One example of such a paradigm is the P300 rapid serial visual presentation (RSVP) speller. In this paradigm letters are presented one after the other in a random order, and the subject is asked to pay attention only to one of the letters called \textit{target} letter or \textit{target stimuli} (by counting them silently, for example). Whenever a subject pays attention to the target letter, a special waveform called P300 is expected to occur when a person's attention is distracted toward a rare event. It is called P300 since there is usually a peak in the EEG amplitude 300ms after the presentation of a target event. The advantage of the RSVP paradigm is that it does not require any eye movements, and can thus be operated by patients who have lost control of their eye gaze completely.

\subsection{``Deep'' Neural Networks - Overview}

Deep Neural Networks (DNNs) are networks constructed from layers of artificial neural networks (ANNs) between the input and output. There are two main type of ANN architectures: Feed Forward Neural Networks (FFNN or FF) and Recurrent Neural Networks (RNN). In FFNN directed cycles are not allowed (i.e., data can flow only to the next layer) while the RNN architecture allows directed cycles within the network (i.e., data can also flow between ``neurons'' in the same layer). The directed cycles in an RNN allows the network to ``remember'' past events, making it suitable for sequence learning~\cite{rumelhart1985learning}.

The architecture we propose is a combination of several ANN layer types described below:

\textbf{Fully Connected Layer - FC} - A layer where each neuron in the input is connected to each neuron in the output is often called Fully Connected Layer or FC. In an FC layer, the output is obtained by the following equation:
\[y=\sigma \left( {xW + b} \right)\]
where $x$ is the input vector, $W$ is a matrix that represent a linear mapping and $b$ is vector that reflects the transformation called the $bias$ ($b$ holds the value for each output unit). $\sigma \left( \cdot \right)$ represent an element-wise non-linearity such as ReLU, sigomid or TanH:

\[{\rm{Relu}}\left( x \right) = \max \left( {x,0} \right)\,\,\,\,\,\,{\rm{sigmoid}}\left( x \right) = \frac{1}{{1 + {e^{ - x}}}}\,\,\,\,\,\,{\mathop{\rm Tan}\nolimits} {\rm{H}}\left( x \right) = \frac{{{e^x} - {e^{ - x}}}}{{{e^x} + {e^{ - z}}}}\]


\textbf{Convolutional Neural Network Layer - CNN }  is a layer that utilizes local correlations between adjacent cells of the input layer. Since, unlike in an FC layer, the output can be of multiple dimensions, we refer to the output as a \textit{Feature Map}. The feature maps are obtained by activating trainable multi-dimension kernels across the input layer. 
Equation \ref{eq:1} describe the output of element $i$ in a 1D CNN layer feature map:

\begin{equation} \label{eq:1}
y\left( i \right) = \sigma \left( {W * x\left[ {i - k,...,i + k} \right] + b} \right)
\end{equation}
If the kernel is 1D filter with a length of $k$ and the next layer has $M$ outputs, then $W$ is a matrix of size $M \times k$.

\begin{figure}[t]
			\centering
			\includegraphics{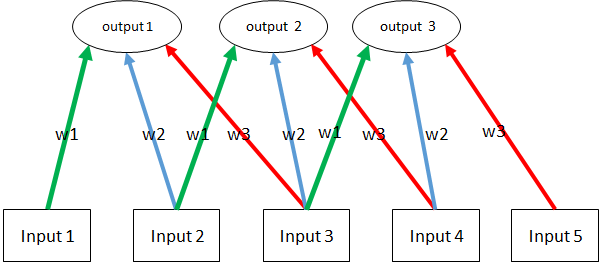}
			\caption[1D CNN layer Schematic diagrams]{1D CNN layer Schematic diagrams}
			\label{fig:CNN_Example}
\end{figure}

\textbf{Recurrent Neural Network Layer - RNN} - The simplest form of an RNN~\cite{rumelhart1985learning, werbos1988generalization} is a layer where the output is connected to the input. Unlike FF layers, in RNN, the result at time $t$ ($y(t)$) is a function of both the current input $x_t$ and the previous state: 

\begin{equation} 
\label{eq:2}
y(t-1)$:$y\left( t \right) = \sigma \left( {x\left( t \right)W + y\left( {t - 1} \right)U + b} \right)
\end{equation}

Here, again $\sigma $ is a non-linear activation function. The structure of an RNN layer, allows the network to contain memory, since it has access to information from previous time-stamps. RNNs are known to suffer from a  phenomenons called "vanishing gradient" and "exploding gradient" \cite{pascanu2013difficulty}: while training, the gradient of the loss function may not propagate to the first layers (i.e., the layers closer to the input layer) or may reach very large values (thus updating the layer weights too much). These problems prevent RNNs from learning long temporal dependencies (See~\cite{bengio1994learning}). A common solution for these problem is called Long Short Term Memory layer.


%
%
%
\textbf{Long Short Term Memory Layer -- LSTM}  is a type of RNN with a special architecture designed to allow overcome RNN difficulties to learn long term dependencies. Each unit in an LSTM layer has two outputs:$y(t)$ - the main output and $c(t)$ - the unit's memory. LSTM uses an architecture with a set of "gates" in order to control the data flow and by that overcome the vanishing and exploding gradient mentioned above. LSTM equations:

\[\begin{array}{l}
f\left( t \right) = {\rm{sigmoid}}\left( {{W_f}x\left( t \right) + {U_f}y\left( {t - 1} \right) + {b_f}} \right)\\
i\left( t \right) = {\rm{sigmoid}}\left( {{W_i}x\left( t \right) + {U_i}y\left( {t - 1} \right) + {b_i}} \right)\\
o\left( t \right) = {\rm{sigmoid}}\left( {{W_o}x\left( t \right) + {U_o}y\left( {t - 1} \right) + {b_o}} \right)\\
c\left( t \right) = f\left( t \right) \circ c\left( {t - 1} \right) + \tanh \left( {i\left( t \right) \circ {W_c}x\left( t \right) + {U_C}y\left( {t - 1} \right) + {b_c}} \right)\\
y\left( t \right) = o\left( t \right) \circ \tanh \left( {c\left( t \right)} \right)
\end{array}\]

${W_{\left\{ {i,p,f,o} \right\}}}$ represent weights of the current state and  ${U_{\left\{ {i,p,f,o} \right\}}}$ represent weights of the previous states.

\subsection{P300 Classification}

There are a lot of methods for identifying the P300 target signal for a BCI task. Blankertz et al.~\cite{P300_Tutorial} suggest to select the time interval with maximal separation between the target and non target samples, average their electro-potential value and use shrinkage LDA to classify these features. Using this method has a drawback due to the low complexity of LDA model \cite{cincotti2003comparison}. The winner of the BCI competition III: dataset II used an ensemble of support vector machines (SVM) \cite{P300SVMWinner}, and other methods include hidden Markov model, k-nearest neighbours, and more  \cite{cincotti2003comparison}.

More recently, given the success of `deep' neural networks \cite{krizhevsky2012imagenet}, there have been several attempts to apply `deep learning' for BCI related tasks. Cecotti and Graser~\cite{P300_CNN} were the first to use CNNs for P300 spelling. In their work, they train an ensemble of CNN-based P300 classifiers to identify the existence of P300. Manor and Geva~\cite{RSVP_P300_geva} used CNN for the RSVP P300 classification task and suggested a new spatio-temporal regularization which have shown improvement in the performance.

Unlike feed forward network models such as CNN and multi-layer perceptron (MLP), the RNN architecture allows directed cycles within the network, which enables the model to ``memorize past events''. LSTM \cite{LSTM_origin} is a type of RNN, which includes a special node that can be described as a differentiable memory cell. The specific architecture of LSTM enables it to overcome some of the weakness of simple RNNs~\cite{bengio1994learning}.

There are several reasons why LSTM is a good candidate for modelling the P300 pattern. First, RNN and LSTM have shown success when modeling time series for tasks such as handwriting and speech recognition \cite{graves2013speech,  graves2008unconstrained, yue2015beyond}. In addition, RNN is known to have the capability to approximate dynamical systems \cite{li2005approximation}, which makes it a natural candidate for modelling the dynamics of EEG data. Another motivation is that RNN can be seen as powerful form of hidden Markov models (HMM), which have been shown to classify EEG successfully \cite{solhjoo2005classification,obermaier2001hidden,cincotti2003comparison}; RNNs can be seen as HMMs with an exponentially large state space and an extremely compact parametrization~\cite{sutskever2009recurrent}.

LSTM was already used for analysing EEG data for emotion detection \cite{soleymani2014continuous} and a phenomena called behavioral microsleeps \cite{davidson2005detecting}. Bahshivan et al.~\cite{LSTM_EEG} modeled inter-subject EEG features for identifying cognitive load by using convolutional LSTM. They created a video from three different band powers in each electrode. One of the major differences between their work and ours is that we use the original signal without any feature extraction (such as band power), and that we focus specifically on P300 speller.

\section{Method}

We compared the performance of LSTM based methods with other methods on a dataset from a RSVP P300 speller study~\cite{BlaknertzExperiment}. We used  average prediction across 10 trials to measure the P300 speller accuracy as applied in~\cite{BlaknertzExperiment}.

\subsection{P300 speller experiments settings}
The dataset includes 55 channels of EEG recordings from 11 subjects. Each subject is presented with 10 repetitions of 60 to 70 sets of 30 different letters and symbols. In total there are approximately 20,000 samples for each subject where 1/30 of them are supposed to contain a P300 target. While the original experiment contains 3 different settings (interval of 116ms with/without colors and 83ms with color), we used the experiment setting of 116ms intervals with letters in different colors. For more detail, see \cite{BlaknertzExperiment}. 

In addition to the filters applied in~\cite{BlaknertzExperiment}, all models that we used share the same pre-processing stage of down-sampling the input frequency from 200Hz to 25 Hz. The result is that each learning sample is a matrix of 55 channels with 25 time samples each, or $55*25 = 1375$ features. Each sample thus covers exactly 1 second around the target event, at times [-200,800] ms.

\subsection{Formulating the BCI task}
In P300 speller the task is to identify to which letter a subject paid his attention to by identifying a P300 pattern in the EEG. This can be done by finding a function $f(x)$ that when given an EEG sample – $x$, return the probability that P300 pattern is found in it. By identifying the EEG sample with the maximal $f(x)$ score among the EEG samples of all the different letters we can identify the target stimuli (i.e. the letter that the subject focuses on). In the following section, the explanation will focus on identifying a single letter and this for the sake of simplifying the explanation.

\subsection{RSVP P300 speller trials}

\paragraph{Notations}
\begin{labeling}{notation}
	\item [$f(x)$] P300 identification function
	\item [$X_c$] An EEG sample data that was recorded when presented character $c$
	\item [$C$] The set of all the available stimulus 
	\item [$\hat{c}$] The predicted target stimuli (i.e. the letter the subject was suppose to focus on)
	\item [$c*$] The true target stimuli (i.e. the letter the subject was suppose to focus on)	
	\item [$R$] Group of trial
	\item [$x_{c,r}$] An EEG sample data that was recorded when presented character $c$ on trial $r$
\end{labeling}

A \underline{\textit{trial}} is the presentation of all the characters in an alphabet, one after the other, in a random order. In each trial, only one letter is the actual target stimuli.  

The predicted character $\hat{c}$ in a single trial is computed by finding the character with maximal value of $f(x)$ among the group of all the letters ($C$):

\begin{equation}
\hat{c} = \arg \max \left\{ {f\left( {{x_{c}}} \right)} \right\}{  _{c \in C}}
\end{equation}

In order to achieve robust character recognition prediction, a common approach is repeating each trial several time and average the prediction for each different stimulus. The group of all repetitions of the same trial will be called $R$. EEG recording where stimuli $c$ presented in trial $r$ is called $x_{c,r}$. The formula for predicting the target stimuli in $R$ attempts is:
\begin{equation}
\hat c = \arg \max \left\{ {\frac{1}{R}\sum\limits_{r \in R}^{} {f\left( {{x_{c,r}}} \right)} } \right\}{_{c \in C}}
\end{equation}

\subsection{Loss function}
In neural network, the weights are updated by deriving the loss function with respected to the network weights. If we train the neural network to identify a P300 target directly (as in \cite{P300_CNN}) the error, depends only on whether the sample contains a P300 target. Here $y_{r,c}$ refer to the true label of sample $x_{r,c}$:
\begin{equation}
E = e\left( {f\left( {{x_{r,c}}} \right),{y_{r,c}}} \right)
\end{equation} 

In this research we use to the binary log loss function:
\begin{equation}\label{eq:binary_log_loss}
e\left( {f\left( {{x_{r,c}}} \right),{y_{r,c}}} \right) =  - \left( {{y_{r,c}}\log \left( {f\left( {{x_{r,c}}} \right)} \right) - \left( {1 - {y_{r,c}}} \right)\log \left( {1 - f\left( {{x_{r,c}}} \right)} \right)} \right)
\end{equation}

The error in a neural network is typically calculated on multiple samples, as follows:
\begin{equation}\label{eq:mini_batch_loss}
E = \frac{1}{M}\sum\nolimits_i^M {e\left( {f\left( {{x_i},{y_i}} \right)} \right)}
\end{equation}

Here $M$ indicates the size of the batch of samples.

\subsection{Classification Models}
\begin{figure*}[t]
		\centering
		\begin{minipage}{.3\textwidth}
			\centering
			\begin{subfigure}{1.0\textwidth}
				\centering
				\includegraphics[height=5cm]{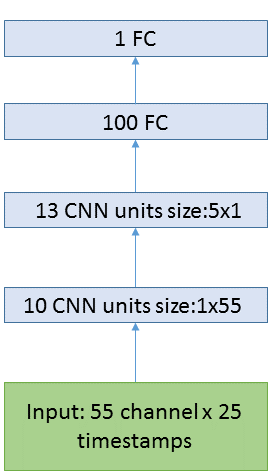}
				\caption{CNN model}
				\label{fig:CNN_model}
			\end{subfigure}
		\end{minipage}	
		\begin{minipage}{.3\textwidth}
			
			\centering
			\begin{subfigure}{1.0\textwidth}
				\centering
				\includegraphics[height=5cm]{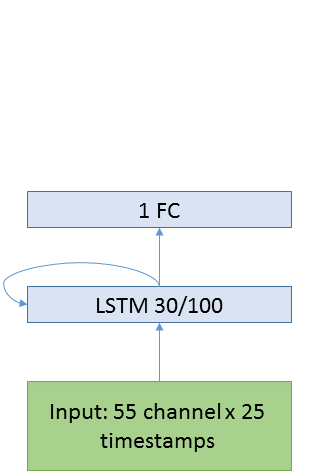}
				\caption{LSTM model}
				\label{fig:LSTM_model}
			\end{subfigure}
			
		\end{minipage}%
		\begin{minipage}{.3\textwidth}
			\centering
			\begin{subfigure}{1.0\textwidth}
			\centering
			\includegraphics[height=5cm]{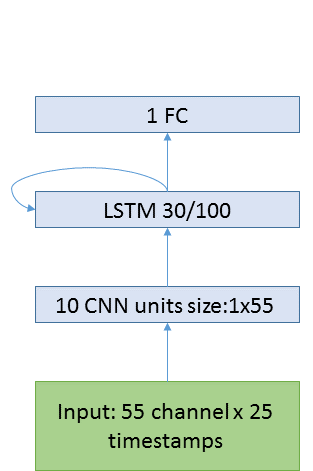}		
			\caption{CNN-LSTM model}
			\label{fig:LSTM_model_CNN}		
			\end{subfigure}
		\end{minipage}
		
		\caption{Schematic diagrams of the neural networks evaluated. FC stands for fully connected layers.}
	\end{figure*}

The models evaluated in this experiment are:
\begin{itemize}
	\item LDA - A common method used in P300 classification for BCI is LDA \cite{BlaknertzExperiment,P300_Tutorial}. Here we will use a simplified version; unlike \cite{BlaknertzExperiment} we use all the timestamps as features, and we are using a non-shrinkage version of LDA.

	\item CNN (Fig.\ref{fig:CNN_model}) -- The CNN model we use is similar to the one used in \cite{P300_CNN}. The first layer is composed of 10  spatial filters, each of size $55*1$ -- the number of channels. The second layer contains 13 different temporal filters with size of $1*5$. Each one of the temporal filters processes 5 subsequent time stamps without overlapping. The third and fourth layers are simple fully connected layers followed by a single cell with sigmoid activation function that emits a scalar.
	
	\item LSTM large/small (Fig.\ref{fig:LSTM_model}) -- LSTM large/small are both composed of single LSTM layers with 100 and 30 hidden cells in each, correspondingly. Both models end with a single cell with sigmoid activation layer that emits a scalar.
	
	\item LSTM-CNN large/small Fig.\ref{fig:LSTM_model_CNN} -- The model has CNN as a first layer (the spatial domain layer) and LSTM as the second layer for the temporal domain. The first convolutional layer is the same as in the CNN model. Unlike the CNN model, the temporal layer is an LSTM layer with 100/30 hidden cells. The last layer contains a single cell a with sigmoid activation layer that emits a scalar.

\end{itemize}

In order to examine the power of each method in modelling the inter-subject and  intra-subject variance we have conducted the following experiments:
\begin{enumerate}
	\item Training and testing on each subject's data separately in order to explore intra-subject generalization.
	\item Training and testing on all the different subjects data combined in order to investigate the impact of larger amounts of data.
	\item Training on all subjects expect one. We conduct this experiment in order to explore the value of using a model that was trained off-line, on different subjects, and then use this model on new subject, with or without additional calibration.
\end{enumerate}
									
A highly desired property from BCI systems is tolerance to a small degree of noise in the stimuli onset time.  In order to evaluate the resistance to such noise, we use a model trained on the original stimuli onset (i.e, noise level = 0ms) and evaluate its performance on different stimuli onset: noise levels of -120ms,-80ms,-40ms, +40ms, +80ms, and 120ms. We conducted this experiment using 10-fold cross validation in order to be able to get statistically significant results. This last experiment was conducted only on the CNN and LSTM-CNN models and used data from all subjects (as in experiment 2 described above).

For all the experiments, the different models were trained using RMSProp~\cite{tieleman2012lecture} optimizer for 30 epochs with a learning rate of 0.001 and then continued to train for 30 epochs with a learning rate of 0.00001.

RMSProp \cite{tieleman2012lecture} is a stochastic gradient descent (SGD) method. Unlike simple SGD, the method can adapt different learning rate for each parameter separately and use moving average across the past gradient in order to scale the learning rate per-feature. We decided to use RMSProp since it is known to be robust and fast \cite{xu2015show, karpathy2015deep, szegedy2016rethinking}.

%
%

\subsection{Implementation and Real Time Possible Usage}
The code was implemented using the Keras framework \cite{chollet2015keras}. Training was conducted using a 4-core i7 laptop with 16Gb RAM. Training took 110 seconds for the small LSTM-CNN model and 24 seconds for the CNN model. The difference is due to the distributed nature of CNN, which allows much of the computation to be computed in parallel. Predicting on single example takes about 0.6 milliseconds. In terms of space, both models require less than 70kb of disk space.

One of the advantages of using ``deep learning'' models is that they allow compressing knowledge from a lot of samples into a compact form. As we show in our experiments, it is possible to pre-train on multiple subjects and then fine tune it to a specific subjects calibration data. For example, training on 3000 calibration samples using the 4-core i7 laptop will take less than a minute (fine-tuning for 30 epochs). After the model is trained, using it for real-time prediction is feasible as well, since predicting each sample takes 0.6 milliseconds. The data and code can be found online\footnote{\url{https://github.com/Ori226/p300_lstm}}. 

\section{Results}

Tab.\ref{table:AllAverageedResults} summarizes the results of the different experiments; all results are based on an average of 10 consecutive trials to detect the target letter, as in~\cite{BlaknertzExperiment}. The results for training and testing on the same subject (Tab.\ref{table:AccuracyPerSubject}) indicates that LSTM is inferior (82\%), and even the LSTM\_CNN combined model performs less than the the simple LDA method (86 and 93\% in the LSTM\_CNN models and 96\% using LDA). A possible advantage for LSTM only becomes apparent with larger amounts of data -- when training and testing on all the subjects together (Tab.\ref{table:AllAverageedResults}). The large LSTM model performs poorly -- 77\%; we suspect it is due to the large number of trainable parameters -- 62501 (``over-fitting''); this is why we introduced CNN as a first layer and reduced number of hidden LSTM cells.

Tab.\ref{table:AccuracyPerSubject} summarizes the results per single subject. When comparing the accuracy result of each subject separately, we can see there is significant difference among subjects, across the different models. For example subject \textit{fat} results in higher accuracy than \textit{icn} regardless of the tested model. Eventually, the best network method -- using training on other subjects and recalibration with a combined CNN-LSTM large model, is able to boost the results of the worse subject to 86\%.

\vspace{0.4cm}	
\begin{table*}[t]
	\captionof{table}{Average accuracy across all experiments.}
	\label{table:AllAverageedResults}
	\centering
	\begin{tabular}{l|ccccc}
		\toprule
		model &  \makecell{number of \\parameter} &  \makecell{accuracy \\
			per subjects} & \makecell{accuracy \\
			all subjects}&\makecell{all but one}&\makecell{all but one \\ after fine tuning}\\
		\midrule
		LDA            &            1375 &                   0.96 &                  0.79 &                           0.65 &                                                 x \\
		LSTM large       &          62501 &                   0.82 &                  0.77 &                              x &                                                 x \\
		LSTM small     &           10351 &                   0.89 &                   0.9 &                              x &                                                 x \\
		CNN            &            7924 &                   0.98 &                  0.92 &                           0.84 &                                              0.97 \\
		\makecell{LSTM-CNN\\large}&           49041 &                   0.93 &                   0.9 &                              x &                                                 x \\
		\makecell{LSTM-CNN\\small}&            5511 &                   0.86 &                  0.93 &                           0.84 &                                              0.97 \\
		\bottomrule
	\end{tabular}
	
\end{table*}

\begin{table*}
	\centering
	\captionof{table}{Average accuracy per subject.}
	\label{table:AccuracyPerSubject}	
	\begin{tabular}{l|ccccccc}
		\toprule
		{subject} &   LDA &   LSTM large &  LSTM-CNN large &   CNN &  LSTM small& LSTM-CNN small \\
		\midrule
		fat     &  1.00 &  0.98 &      0.98 &  0.98 &        1.00 &            0.95 \\
		gcb     &  0.91 &  0.82 &      0.88 &  0.92 &        0.74 &            0.75 \\
		gcc     &  1.00 &  0.84 &      0.92 &  1.00 &        0.92 &            0.97 \\
		gcd     &  0.97 &  0.80 &      0.90 &  1.00 &        0.76 &            0.93 \\
		gcf     &  1.00 &  0.92 &      0.94 &  0.95 &        0.97 &            0.95 \\
		gcg     &  0.94 &  0.74 &      0.96 &  0.96 &        0.80 &            0.87 \\
		gch     &  0.97 &  0.93 &      0.96 &  0.97 &        0.97 &            0.96 \\
		iay     &  0.94 &  0.62 &      0.92 &  0.98 &        0.75 &            0.86 \\
		icn     &  0.94 &  0.62 &      0.86 &  0.98 &        0.77 &            0.77 \\
		icr     &  0.93 &  0.97 &      0.98 &  0.98 &        0.98 &            0.98 \\
		pia     &  0.97 &  0.82 &      0.94 &  1.00 &        0.77 &            0.81 \\
		mean    &  0.96 &  0.82 &      0.93 &  0.98 &        0.86 &            0.89 \\
		\bottomrule
	\end{tabular}
	
\end{table*}

In the second stage, we continue training the model on the rest $3/4$ of the \textbf{test subject's} data using a smaller learning rate (0.0001 using RMSProp) for 30 epochs. The second training stage results are presented in columns \textit{CNN and LSTM-CNN all except one fine tune}. The results indicate that as in the other cross-subject evaluation, the LDA accuracy is much poorer than those of the CNN and LSTM-CNN models (65\% as opposed to 84\%). When we allow calibrating the model for each subject, we achieve an average accuracy of 97\% for both CNN and LSTM-CNN.

\begin{table*}[t]
	\centering
	\captionof{table}{Accuracy when training and testing on different subjects. }
	\label{table:AllExceptOne}	
	\begin{tabular}{l|ccccc}
		\toprule
		{subject} & \makecell{LDA \\ all except one} &  \makecell{CNN\\all except one}&  \makecell{CNN\\all except one\\fine tune}& \makecell{SMALL LSTM-CNN\\all except one}&\makecell{SMALL LSTM-CNN\\all except one \\ fine tune} \\
		\midrule
		fat     &                  0.94 &                  1.00 &                            1.00 &                             0.98 &                                               1.00 \\
		gcb     &                  0.43 &                  0.83 &                            0.91 &                             0.86 &                                               0.92 \\
		gcc     &                  0.79 &                  0.98 &                            0.98 &                             0.95 &                                               0.97 \\
		gcd     &                  0.66 &                  0.80 &                            0.99 &                             0.83 &                                               0.97 \\
		gcf     &                  0.68 &                  0.89 &                            0.98 &                             0.79 &                                               0.98 \\
		gcg     &                  0.52 &                  0.81 &                            0.94 &                             0.77 &                                               0.90 \\
		gch     &                  0.87 &                  0.97 &                            0.97 &                             0.97 &                                               0.99 \\
		iay     &                  0.48 &                  0.69 &                            0.98 &                             0.67 &                                               0.97 \\
		icn     &                  0.44 &                  0.58 &                            0.92 &                             0.61 &                                               0.95 \\
		icr     &                  0.63 &                  0.81 &                            1.00 &                             0.89 &                                               1.00 \\
		pia     &                  0.77 &                  0.87 &                            0.96 &                             0.91 &                                               0.97 \\
		mean    &                  0.65 &                  0.84 &                            0.97 &                             0.84 &                                               0.97 \\
		\bottomrule
	\end{tabular}
\end{table*}

\vspace{5mm}

Resistance to temporal noise is displayed in Tab.\ref{table:ResistenceToNoise}. LSTM-CNN shows a significant advantage over both LDA and CNN when testing with stimuli onset different than the one used for training. LSTM-CNN-small achieves an accuracy higher by 3\% and 6\% when adding or removing 40ms to the original stimuli onset, and a t-test indicates that the difference between each pair of groups are statistically significant ($p < 0.05$ - marked in bold). LDA accuracy fall by more than 20\% when facing temporal noise.

\begin{figure*}
	\centering
	\begin{subfigure}{.5\textwidth}
		\centering
		\captionsetup{justification=raggedright,
			singlelinecheck=false,
			margin=9em
		}
		\includegraphics[width=0.9\textwidth]{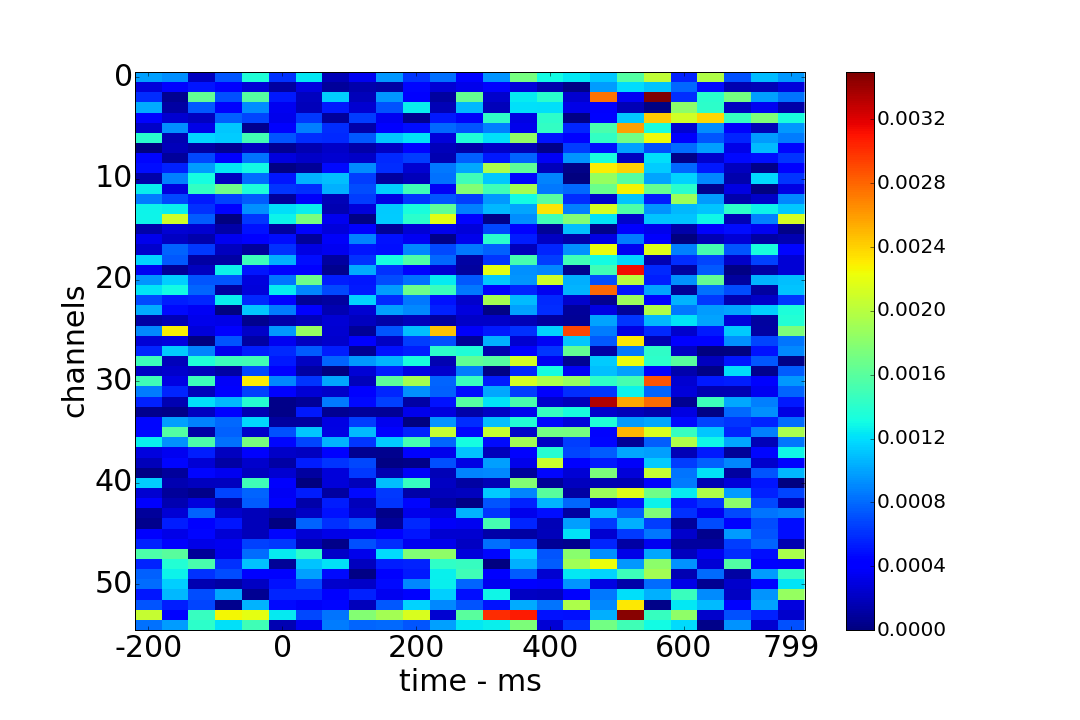}
		\caption{CNN}
		\label{fig:CNNSaliencyMap2}
	\end{subfigure}\hspace*{-3em}
	\begin{subfigure}{.5\textwidth}
		\captionsetup{justification=raggedright,
			singlelinecheck=false,
			margin=7em
		}
		\centering
		\includegraphics[width=0.9\textwidth,left]{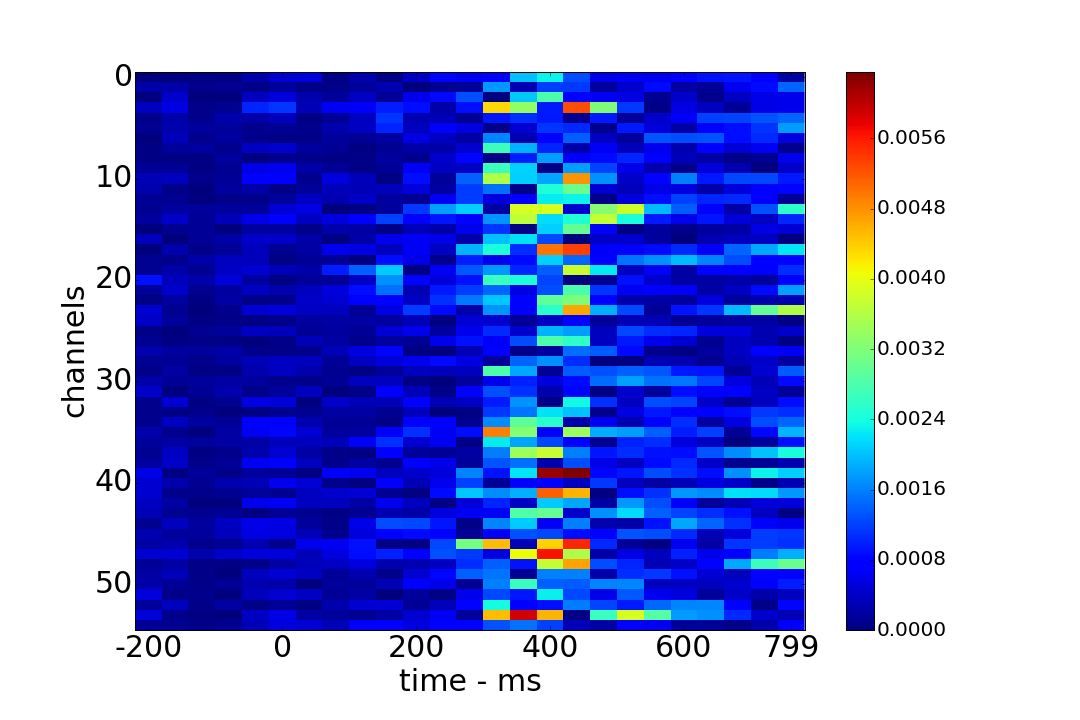}
		\centering
		\caption{LSTM-CNN}
		\label{fig:LSTMCNNSaliencyMap2}
	\end{subfigure}
	\caption{Average gradient across target samples.}
	\label{fig:t}
\end{figure*}

A possible explanation can be seen when looking at the two models' saliency map (Fig.\ref{fig:t}). In order to investigate the ``attention'', or the sensitivity of the LSTM model, and compare it to the CNN model, we used a technique suggested by~\cite{graves2012supervised} and draw the absolute gradient of the neural network with respect to the input.

If $f(x_{1}, ..., x_{n})$ is a differentiable, scalar-valued function, its gradient is the vector whose components are the $n$ partial derivatives of $f$, which is a vector-valued function. In our case $f(x|\theta)$ is the neural network with fixed weights $\theta$ and input $x$. The partial derivatives of $f(x|\theta)$ with respect to $x$ can be interpreted as ``how changing each value of $x$ will change the prediction score''. This gradient should not be confused with the gradient used for training, where the goal is to optimize the model parameters $\theta$ when $x$ is fixed.

	In the case of P300 prediction, $x$ is a matrix of $C\times{T}$ ($C$ - number of channels, $T$ - number of time steps) and $f(x|\theta)$ is the neural network where $\theta$ is the model's weights after training. The gradient $\nabla{f(x|\theta)}$ (see Eq.\ref{eq:4}) is a matrix with the same size as the input $x$, where the amplitude of each cell reflects its impact on the function value. Cells with high absolute value can be interpreted as the cells that have a significant influence on the prediction function.
	
	\begin{equation}\label{eq:4}
	\nabla f\left( {x|\theta } \right) = \left[ {\begin{array}{*{20}{c}}
		{\frac{{\partial f\left( {x|\theta } \right)}}{{\partial x\left( {{c_1},{t_1}} \right)}}}&{...}&{\frac{{\partial f\left( {x|\theta } \right)}}{{\partial x\left( {{c_1},{t_T}} \right)}}}\\
		{...}&{...}&{...}\\
		{\frac{{\partial f\left( {x|\theta } \right)}}{{\partial x\left( {{c_C},{t_1}} \right)}}}&{...}&{\frac{{\partial f\left( {x|\theta } \right)}}{{\partial x\left( {{c_C},{t_T}} \right)}}}
		\end{array}} \right]
	\end{equation}

The results displayed in Fig.\ref{fig:CNNSaliencyMap2} and Fig.\ref{fig:LSTMCNNSaliencyMap2} show the average absolute gradient across all the \textit{target} samples of a single cross validation test data: the warm colors correspond to high gradient values, indicating that the model is more sensitive to change in this input feature. We can see the sensitivity of the CNN model spread across the recording relatively evenly as opposed to the LSTM-CNN which is focused around the 250ms and 450ms time-stamps.

\begin{small}
	\captionof{table}{Accuracy when introducing temporal noise.}
	\label{table:ResistenceToNoise}
	\centering
	
	\begin{tabular}{l|ccc}
		\toprule
		{Noise} &  CNN &  LSTM\_CNN &  LDA\\
		\midrule
		-120  &         0.058 &              0.044 &         0.016 \\
		-80   &         0.275 &              0.299 &         0.016 \\
		-40   &         \textbf{0.825} &              \textbf{0.864} &  \textbf{0.565} \\
		40   &         \textbf{0.848} &              \textbf{0.896} &   \textbf{0.608} \\
		80   &         0.335 &              0.390 &         0.260 \\
		120  &         0.042 &              0.042 &         0.059 \\
		\bottomrule
	\end{tabular}
	
\end{small}

\vspace{0.5cm}

%

\section{Discussion and Future Direction}
In this work we examined using LSTM neural networks for the task of the BCI task of P300 speller. Despite its temporal nature, no version of LSTM investigated in this work has shown a significant advantage compared to the CNN model suggested by \cite{P300_CNN}. We did see LSTM results improve with large amounts of data from multiple subjects, and superior results with a combined CNN-LSTM model; moreover, we have shown that this combined model is significantly more robust to temporal noise in the stimuli onset. We also show that the sensitivity of the LSTM based model is much more focused on the area between 250ms to 450ms than CNN based model. This sensitivity is correlated with what we know about the P300 ERP (a peak around 300ms after the stimuli onset). We thus believe that the smaller area of the sensitivity explains the robustness of the LSTM model to noise in the time domain, since it is less sensitive to the data outside the P300 phenomena. 

In our research we have used one of the largest EEG datasets for supervised machine learning -- approximately 20,000 labelled samples per subject. Nevertheless, for a single subject we do not see any advantage in using ''deep learning'' over simple linear methods such as LDA. This is in contrast to other reports, e.g.~\cite{LSTM_EEG} report an improvement of 8.9\% to 15.3\% (reduction of error) on a working memory dataset with 2670 labelled samples. The benefits of ``deep learning'' models can be seen in transfer learning among subjects. We did not find evidence that RNNs may be superior to CNNs in classifying EEG patterns, although the LSTM model was more robust to noise. 
Each sample in our experiment had a fixed length, which, allowed us to use the feed forward models such as CNN. Future work on RNN may investigate using sequence-to-sequence training with variable length samples (such as suggested by \cite{graves2012supervised}), where RNNs have advantage over the feed forward models.





\section{References}
\bibliography{bci_conf}

\begin{thebibliography}{10}

\bibitem{BlaknertzExperiment}
Laura Acqualagna and Benjamin Blankertz.
\newblock Gaze-independent {BCI}-spelling using rapid serial visual
  presentation (rsvp).
\newblock {\em NeuroImage}, 124:901--908, 2013.

\bibitem{LSTM_EEG}
et~al. Bashivan, Pouya.
\newblock Learning representations from eeg with deep recurrent-convolutional
  neural networks.
\newblock {\em arXiv}, 1511.06448, 2015.

\bibitem{bengio1994learning}
Yoshua Bengio, Patrice Simard, and Paolo Frasconi.
\newblock Learning long-term dependencies with gradient descent is difficult.
\newblock {\em IEEE transactions on neural networks}, 5(2):157--166, 1994.

\bibitem{P300_Tutorial}
et~al. Blankertz, Benjamin.
\newblock Single-trial analysis and classification of erp components—a
  tutorial.
\newblock {\em NeuroImage}, 50:814--825, 2011.

\bibitem{P300_CNN}
Hubert Cecotti and Axel Graser.
\newblock Convolutional neural networks for p300 detection with application to
  brain-computer interfaces.
\newblock {\em IEEE transactions on pattern analysis and machine intelligence},
  33:433--445, 2011.

\bibitem{chollet2015keras}
Fran\c{c}ois Chollet et~al.
\newblock Keras.
\newblock \url{https://github.com/fchollet/keras}, 2015.

\bibitem{cincotti2003comparison}
Febo Cincotti, A~Scipione, A~Timperi, D~Mattia, AG~Marciani, J~Millan,
  S~Salinari, L~Bianchi, and F~Bablioni.
\newblock Comparison of different feature classifiers for brain computer
  interfaces.
\newblock pages 645--647, 2003.

\bibitem{davidson2005detecting}
PR~Davidson, RD~Jones, and MTR Peiris.
\newblock Detecting behavioural microsleeps using eeg and lstm recurrent neural
  networks.
\newblock 27:5754--5757, 2005.

\bibitem{FirstP300}
Lawrence~Ashley Farwell and Emanuel Donchin.
\newblock Talking off the top of your head: toward a mental prosthesis
  utilizing event-related brain potentials.
\newblock {\em Neural Computation}, 70:510--523, 1988.

\bibitem{graves2012supervised}
Alex Graves.
\newblock Supervised sequence labelling.
\newblock pages 5--13, 2012.

\bibitem{graves2008unconstrained}
Alex Graves, Marcus Liwicki, Horst Bunke, J{\"u}rgen Schmidhuber, and Santiago
  Fern{\'a}ndez.
\newblock Unconstrained on-line handwriting recognition with recurrent neural
  networks.
\newblock pages 577--584, 2008.

\bibitem{graves2013speech}
Alex Graves, Abdel-rahman Mohamed, and Geoffrey Hinton.
\newblock Speech recognition with deep recurrent neural networks.
\newblock pages 6645--6649, 2013.

\bibitem{hinton2012deep}
Geoffrey Hinton, Li~Deng, Dong Yu, George~E Dahl, Abdel-rahman Mohamed, Navdeep
  Jaitly, Andrew Senior, Vincent Vanhoucke, Patrick Nguyen, Tara~N Sainath,
  et~al.
\newblock Deep neural networks for acoustic modeling in speech recognition: The
  shared views of four research groups.
\newblock {\em IEEE Signal Processing Magazine}, 29(6):82--97, 2012.

\bibitem{LSTM_origin}
Sepp Hochreiter and Jürgen Schmidhuber.
\newblock Long short-term memory.
\newblock {\em arXiv}, 9.8:1735--1780, 1997.

\bibitem{karpathy2015deep}
Andrej Karpathy and Li~Fei-Fei.
\newblock Deep visual-semantic alignments for generating image descriptions.
\newblock In {\em Proceedings of the IEEE Conference on Computer Vision and
  Pattern Recognition}, pages 3128--3137, 2015.

\bibitem{krizhevsky2012imagenet}
Alex Krizhevsky, Ilya Sutskever, and Geoffrey~E Hinton.
\newblock Imagenet classification with deep convolutional neural networks.
\newblock pages 1097--1105, 2012.

\bibitem{Lenet98}
et~al. LeCun, Yann.
\newblock Gradient-based learning applied to document recognition.
\newblock {\em Proceedings of the IEEE}, 86:2278--2324, 1998.

\bibitem{li2005approximation}
Xiao-Dong Li, John~KL Ho, and Tommy~WS Chow.
\newblock Approximation of dynamical time-variant systems by continuous-time
  recurrent neural networks.
\newblock {\em IEEE Transactions on Circuits and Systems II: Express Briefs},
  52(10):656--660, 2005.

\bibitem{lotte2007review}
Fabien Lotte, Marco Congedo, Anatole L{\'e}cuyer, Fabrice Lamarche, and Bruno
  Arnaldi.
\newblock A review of classification algorithms for eeg-based brain--computer
  interfaces.
\newblock {\em Journal of neural engineering}, 4(2):R1, 2007.

\bibitem{RSVP_P300_geva}
Ran Manor and Amir~B. Geva.
\newblock Convolutional neural network for multi-category rapid serial visual
  presentation{BCI}.
\newblock {\em Frontiers in computational neuroscience}, 9, 2015.

\bibitem{obermaier2001hidden}
Bernhard Obermaier, Christoph Guger, Christa Neuper, and Gert Pfurtscheller.
\newblock Hidden markov models for online classification of single trial eeg
  data.
\newblock {\em Pattern recognition letters}, 22(12):1299--1309, 2001.

\bibitem{pascanu2013difficulty}
Razvan Pascanu, Tomas Mikolov, and Yoshua Bengio.
\newblock On the difficulty of training recurrent neural networks.
\newblock {\em ICML (3)}, 28:1310--1318, 2013.

\bibitem{P300SVMWinner}
Alain Rakotomamonjy and Vincent Guigue.
\newblock {BCI} competition iii: dataset ii-ensemble of svms for {BCI} p300
  speller.
\newblock {\em IEEE transactions on biomedical engineering}, 55(3):1147--1154,
  2008.

\bibitem{rumelhart1985learning}
David~E Rumelhart, Geoffrey~E Hinton, and Ronald~J Williams.
\newblock Learning internal representations by error propagation.
\newblock Technical report, DTIC Document, 1985.

\bibitem{soleymani2014continuous}
Mohammad Soleymani, Sadjad Asghari-Esfeden, Maja Pantic, and Yun Fu.
\newblock Continuous emotion detection using eeg signals and facial
  expressions.
\newblock pages 1--6, 2014.

\bibitem{solhjoo2005classification}
Soroosh Solhjoo, Ali~Motie Nasrabadi, and Mohammad Reza~Hashemi Golpayegani.
\newblock Classification of chaotic signals using hmm classifiers: Eeg-based
  mental task classification.
\newblock pages 1--4, 2005.

\bibitem{sutskever2009recurrent}
Ilya Sutskever, Geoffrey~E Hinton, and Graham~W Taylor.
\newblock The recurrent temporal restricted boltzmann machine.
\newblock pages 1601--1608, 2009.

\bibitem{szegedy2016rethinking}
Christian Szegedy, Vincent Vanhoucke, Sergey Ioffe, Jon Shlens, and Zbigniew
  Wojna.
\newblock Rethinking the inception architecture for computer vision.
\newblock In {\em Proceedings of the IEEE Conference on Computer Vision and
  Pattern Recognition}, pages 2818--2826, 2016.

\bibitem{tieleman2012lecture}
Tijmen Tieleman and Geoffrey Hinton.
\newblock Lecture 6.5-rmsprop: Divide the gradient by a running average of its
  recent magnitude.
\newblock {\em COURSERA: Neural networks for machine learning}, 4(2), 2012.

\bibitem{werbos1988generalization}
Paul~J Werbos.
\newblock Generalization of backpropagation with application to a recurrent gas
  market model.
\newblock {\em Neural networks}, 1(4):339--356, 1988.

\bibitem{xu2015show}
Kelvin Xu, Jimmy Ba, Ryan Kiros, Kyunghyun Cho, Aaron~C Courville, Ruslan
  Salakhutdinov, Richard~S Zemel, and Yoshua Bengio.
\newblock Show, attend and tell: Neural image caption generation with visual
  attention.
\newblock In {\em ICML}, volume~14, pages 77--81, 2015.

\bibitem{yue2015beyond}
Joe Yue-Hei~Ng, Matthew Hausknecht, Sudheendra Vijayanarasimhan, Oriol Vinyals,
  Rajat Monga, and George Toderici.
\newblock Beyond short snippets: Deep networks for video classification.
\newblock pages 4694--4702, 2015.

\end{thebibliography}
\bibliographystyle{plain}

\end{document}